\documentclass[journal, compsoc]{IEEEtran} 
\usepackage{graphicx}
\usepackage[utf8]{inputenc}
\usepackage{hyperref} 
\usepackage{amssymb} 
\usepackage{amsmath} 
\usepackage{caption}
\usepackage{subcaption}
\usepackage{natbib}
\usepackage{siunitx} 
\usepackage{tabularx}
\setcitestyle{authoryear}

\makeatletter
\def\@seccntformatinl#1{\csname the#1dis\endcsname\hskip 1em\relax}
\makeatother

\begin{document}
%
\title{A procedure for automated tree pruning suggestion using LiDAR scans of fruit trees.}

\author{Fred~Westling*,
    James~Underwood,
    Mitch~Bryson
    \thanks{All authors are with the University of Sydney}
    \thanks{*Correspondence: f.westling@acfr.usyd.edu.au}
}

\maketitle

\begin{abstract}
    In fruit tree growth, pruning is an important management practice for preventing overcrowding, improving canopy access to light and promoting regrowth.
    In fruit with a high energy content, including avocado (Persea americana), ensuring all parts of the canopy have sufficient exposure to light is of particular importance.
    Due to the slow nature of agriculture and the numerous parameters contributing to yield, decisions in pruning, particularly in selective limb removal, are typically made using tradition or rules of thumb rather than data-driven analysis.
    Many existing algorithmic, simulation-based approaches rely on high-fidelity digital captures or purely computer-generated fruit trees, and are unable to provide specific results on an orchard scale.
    We present a framework for suggesting pruning strategies on LiDAR-scanned commercial fruit trees using a scoring function with a focus on improving light distribution throughout the canopy.
    Due to the destructive nature of physical experimentation, this framework is presented using a three-stage approach where stages can be independently validated.
    Firstly, a scoring function to assess the quality of the tree shape based on its light availability and size was developed for comparative analysis between trees using observations from agricultural literature, and was validated against yield characteristics from an avocado and mango orchard.  This demonstrated a reasonable correlation against fruit count, with an $R^2$ score of 0.615 for avocado and 0.506 for mango.
    The second stage was to implement a tool for simulating pruning by algorithmically estimating which parts of a tree point cloud would be removed given specific cut points using structural analysis of the tree.  This was validated experimentally using manually generated ground truth pruned tree models, showing good results with an average F1 score of 0.78 across 144 experiments.
    Finally, new pruning locations were suggested by discovering points in the tree which negatively impact the light distribution, and we used the previous two stages to estimate the improvement of the tree given these suggestions.  These results were compared to a tree which was commercially pruned using existing wisdom.  The light distribution was improved by up to 25.15\%, demonstrating a 16\% improvement over the commercial pruning, and certain cut points were discovered which improved light distribution with a smaller negative impact on tree volume.
    The final results suggest value in the framework as a decision making tool for commercial growers, or as a starting point for automated pruning since the entire process can be performed with little human intervention.
    Further development should be performed to improve the suggestion mechanism and incorporate more agricultural objectives and operations.
\end{abstract}


\IEEEpeerreviewmaketitle

\section{Introduction}
\label{sec:introduction}

Tree pruning is an important part of fruit tree orchard management, but must be done carefully.  \cite{miller1960avocado} demonstrated that severe pruning can reduce the yield of the current crop, but increase it in following seasons.  As such, it can be a difficult process to inform and feed back for improved decisions.
Since tree growth takes many years, there is interest in performing tree analysis in silico to judge decision making or allow study of physically challenging or destructive operations.
In this work we focus primarily on avocado and mango trees, though methods developed are based on research from a variety of fruit, and should be applicable to any fruit for which light availability is important in growth.

\begin{figure}[ht]
    \centering
    \begin{subfigure}[t]{0.5\columnwidth}
        \centering
        \includegraphics[width=\textwidth]{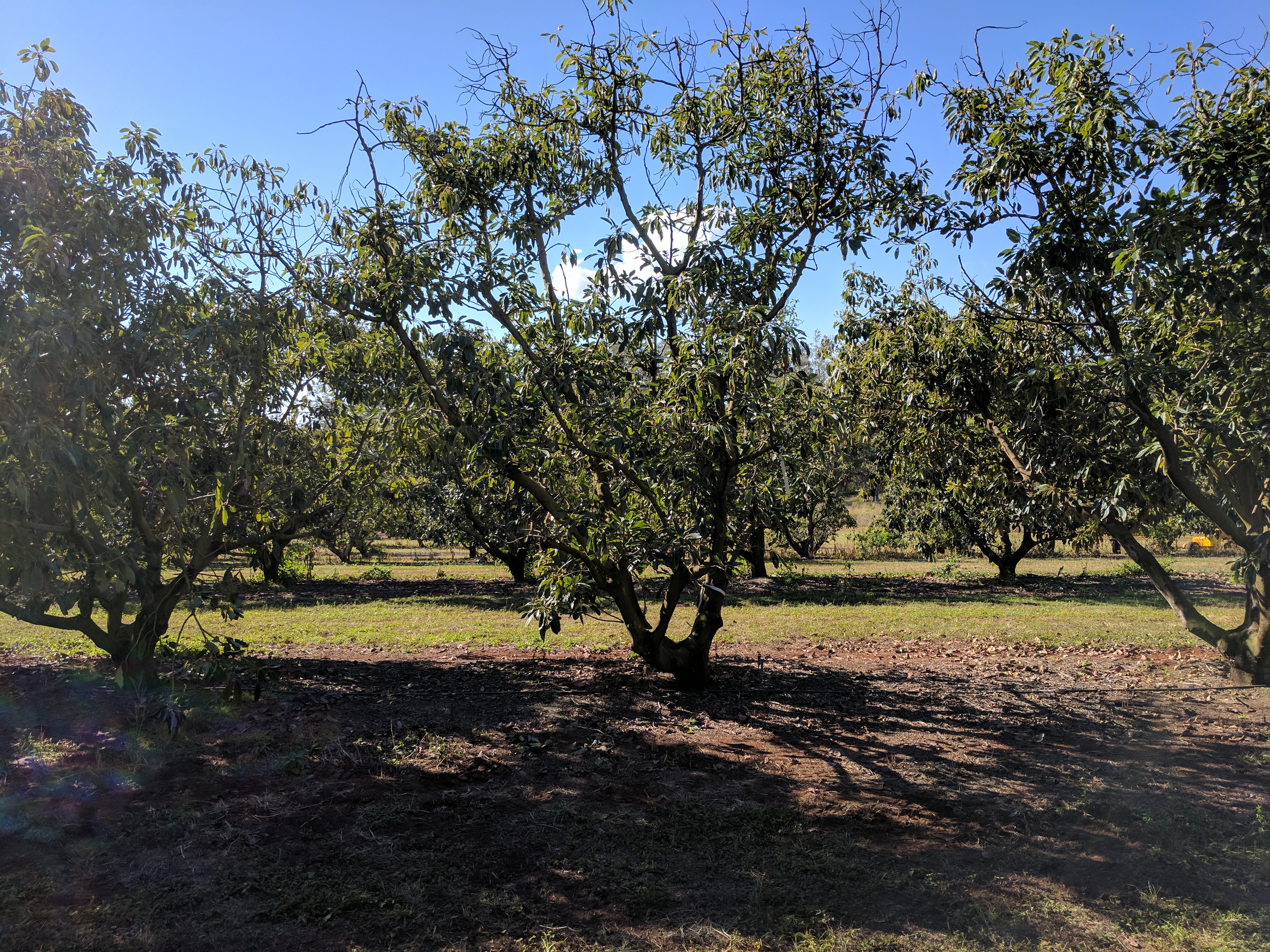}
        \caption{Unpruned, camera}
    \end{subfigure}~
    \begin{subfigure}[t]{0.5\columnwidth}
        \centering
        \includegraphics[width=\textwidth]{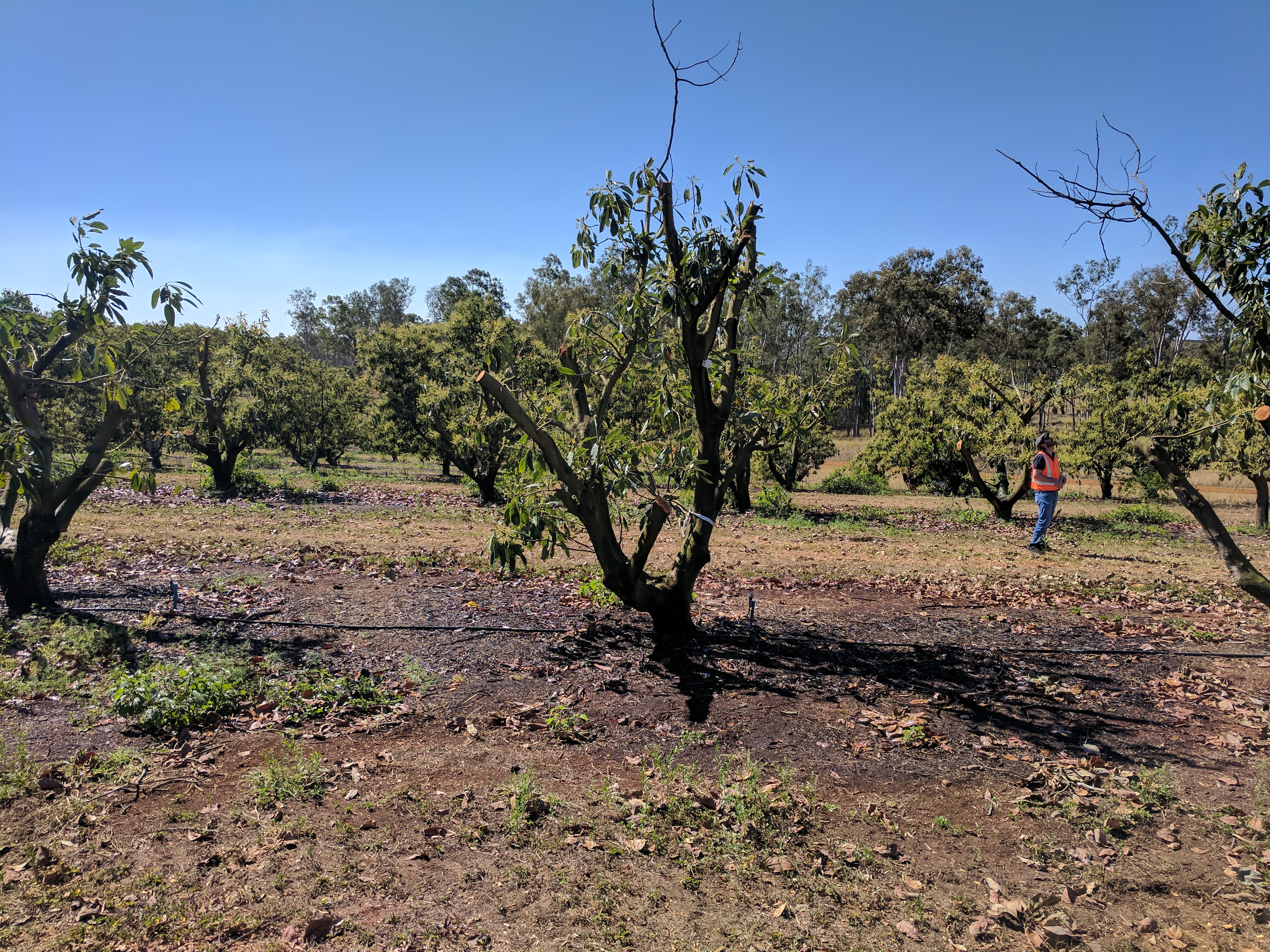}
        \caption{Pruned, camera}
    \end{subfigure}
    \begin{subfigure}[t]{0.5\columnwidth}
        \centering
        \includegraphics[width=\textwidth]{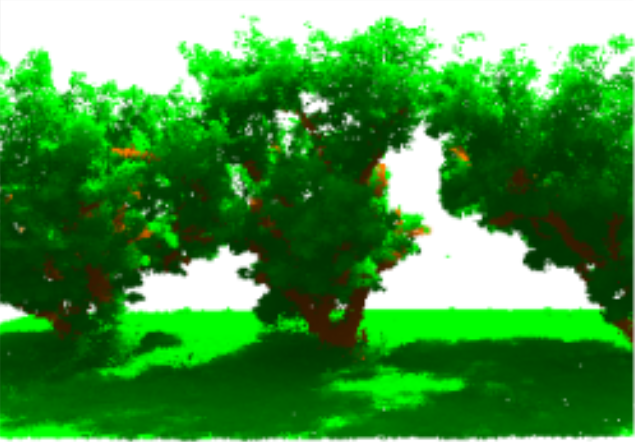}
        \caption{Unpruned, labelled LiDAR}
    \end{subfigure}~
    \begin{subfigure}[t]{0.5\columnwidth}
        \centering
        \includegraphics[width=\textwidth]{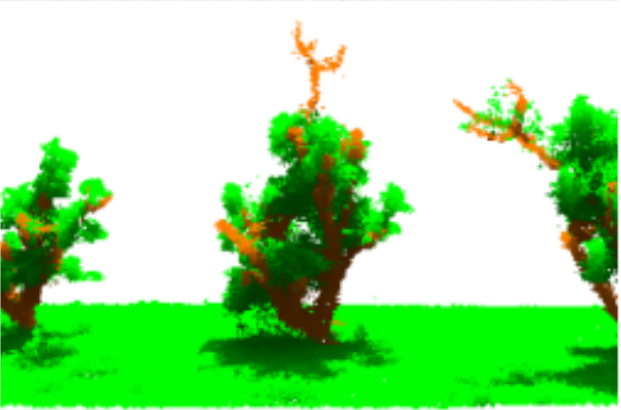}
        \caption{Pruned, labelled LiDAR}
    \end{subfigure}
    \caption{Example of pruning applied to an avocado tree.  The LiDAR scans were captured at the same time the photos were, and were then manually labelled with trunk and foliage and raytraced with simulated light for clarity.}
    \label{fig:real-prune}
\end{figure}

Pruning can be undertaken for a variety of reasons, including lowering tree height to reduce harvesting costs and reduce mutual shading of scaffold branches (\cite{thorp2001pruning}) as well as reducing inter-tree crowding, promoting regrowth and risk mitigation (\cite{purcell2015tree}).
\cite{partida1996avocado} observed that pruning 'Hass' avocado trees produced healthier trees and grew larger sized fruit, and that maintaining the trees at a lower height increased overall tree vitality and fruit production compared to trees which were allowed to grow at will.
Beyond individual tree health, pruning also improves productivity by reducing crowding.
\cite{wilkie2019relationships} showed that, in conventional large-tree low-density orchards, the yield/ha increases with total light interception and canopy volume up to a certain point, and then declines.
In avocado trees in particular, the two primary methods of pruning are "hedging" (planar cuts) and "selective limb removal".
\cite{menzel2014increasing} noted that hedging is most effective in warm subtropical coastal areas, while limb removal is more effective in cool temperate areas where successive crops overlap.
Figure~\ref{fig:real-prune} shows an example of conventional avocado trees which have undergone both processes, significantly reducing their size and quantity of leaf matter to stimulate future yields.


Given the slow nature of growing studies and the interdependent variables contributing to yield, pruning recommendations are typically drawn from old practices or studies in similar tree crops.
Our focus is on pruning decisions designed to improve light penetration through the canopy, which \cite{stassen1999results} lists as a key consideration in fruit production.
For avocado trees in particular, \cite{mickelbart2004sun}notes that the leaves respond relatively slowly to light, so increasing access to sunlight through pruning should improve fruit quality and yield.
\cite{snijder1995strategies} showed that light penetration within the canopy can be improved from 7\% to 58\% through selective pruning, while \cite{stassen1999results} notes that maximum photosynthesis occurs at 30\% or more of full sunlight intensity.
This was further validated by the findings of \cite{marini2020physiology}, who showed that peach and other high-sugar fruit require all areas of the tree to achieve 25\% of full sunlight for optimum growth.
The decisions used to open up the canopy tend to be ad-hoc, though care must be taken to balance other pruning considerations.
For instance, \cite{Orn2016} found a correlation between total light and fruit quality, but also a strong link between volume and yield.
\cite{thorp2001pruning} demonstrated that tree height could be reduced to 6m without negatively impacting yield, but not to 4m.
In general, fruit tree pruning is a process of identifying specific goals and shaping the tree to achieve them while minimising negative effects.

Functional-structural plant modelling (\cite{White2012,White2016}) has been used to generate virtual trees using algorithmic growth to study parameters including light, nutrients and spacing.
\cite{da2014light} and \cite{tang2015light} generate computer models of apple and peach trees respectively to study light interception efficiency, while \cite{yang2016canopy} used similar models to analyse canopy structure.
Others have used captures of real trees rather than purely virtual ones.
\cite{sinoquet2007using} digitised trees in high quality using a magnetic device to accurately recreate leaf angle distributions and plant components to correlate fruit-scale light interception with fruit sugar content.  Methods like these tend to be difficult to perform at scale, while \cite{westling2018light} used fast hand-held laser scanners to generate lower-quality tree models which nevertheless allowed light interception and distribution studies using public weather data.

Physical pruning is destructive by its very nature so virtual approaches to analysing or predicting its effects can be helpful, though few works have explored this space.
By comparing point clouds at different times, \cite{xiao2012change} perform change detection of trees in urban areas using aerial scans.
Similarly, \cite{Estornell2015} presented an approach to estimate the amount of biomass removed during pruning.
\cite{tagarakis2018evaluation} showed the potential to use LiDAR sensors for scanning olive trees prior to pruning to help inform management practices.
Recently, \cite{strnad2020framework} presented a framework for optimising virtual tree pruning against multiple objectives by simulating the effect of pruning and the subsequent growth of new matter.  However, their framework operates purely on virtual trees which are fully observed.

If tree pruning were to be automated, it must be possible to make pruning decisions autonomously, in particular the limb removal decisions which are specific to individual trees that must be observed.
Mobile scanning platforms like that presented by \cite{underwood2016mapping} enable easy and automated 3D point cloud capture of orchard trees, while the method of \cite{westling2018light} allows rapid analysis of the light interception and distribution of a tree using low-quality point clouds.
Here we present a method for non-destructive analysis of the impact of pruning decisions, specifically on the immediate light distribution characteristics on a 3D point cloud model of individual trees as captured by a handheld laser scanner.
Due to the difficulty of capturing real-world data during pruning and tracking commercial pruning decisions, we develop a virtual tree data set using the SimTreeLS system presented by \cite{westling2020simtreels} which can be used to produce realistic LiDAR data upon which pruning analysis can be conducted, though we demonstrate the ability to transfer the method to real data.

\section{Method}
\label{sec:method}

Real-world LiDAR point cloud data were captured at Simpson's avocado farm, a commercial orchard in Childers, Queensland.
Avocado trees are well structured for pruning, with wide splits and long, curving branches.
In 2017 and 2019, we captured data before all pruning and after pruning.  In 2017 in particular we were able to capture before hedging, before limb removal and after all pruning, though we were not able to replicate this in 2019 due to operational constraints.
The 2017 dataset has been made public (\cite{westling2021avocado2}).

Prior to the processes described in this section, we isolate each tree using ground detection and the segmentation procedure presented by \cite{westling2020graph}.
Separately, we generate simulated LiDAR scans of trees using \cite{westling2020simtreels} for ground truth analysis of our pruning simulator, so that we could know exactly where the cut points are and what effect it had on the tree.
To acquire perfectly labelled pruning data for ground truth, we manually "pruned" several trees at the mesh stage and then simulate the LiDAR scan so we have virtual scans before and after pruning for between 1 and 6 limbs removed.

The operations developed here were primarily using the ACFR Comma and Snark open-source libraries (\cite{acfrcomma}) and have been published to \url{https://github.com/fwestling/GraphTreeLS}.

\subsection{Scoring function}
\label{sec:method-scoring}

\begin{figure}[ht]
    \centering
    \includegraphics[width=\columnwidth]{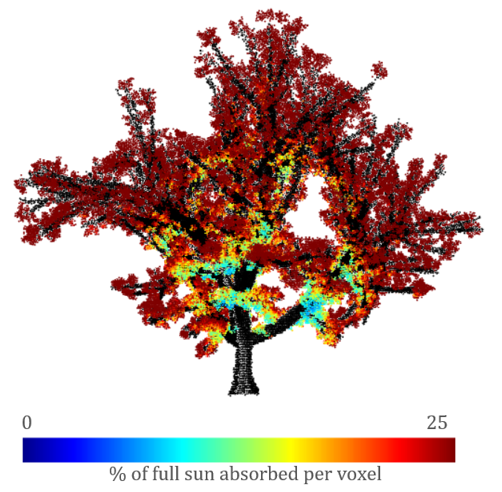}
    \caption{Point cloud of a tree, coloured by light distribution statistic.  Trunks points (perfectly known since this is a simulated tree) are coloured in black.  Voxels with less than 25\% of full sun absorbed are considered suboptimal for production of high quality fruit.}
    \label{fig:d-score-demo}
\end{figure}

In order to compare tree quality, we developed a scoring function based on the literature presented in Section~\ref{sec:introduction}.
Specifically, \cite{marini2020physiology} and \cite{stassen1999results} showed that ideal light distribution for fruit trees occurs when all areas of the tree have access to 25-30\% of available sunlight, and \cite{Orn2016} found good correlations between fruit yield, tree volume and total light absorbed.
In order to balance these considerations, we propose a scoring function which promotes total light absorbed and tree volume while penalising leafy matter with access to less than 25\% of full sun.
Figure~\ref{fig:d-score-demo} shows a point cloud of a tree in which each voxel is coloured by the percentage of light absorbed, with all blue and yellow voxels demonstrating a sub-optimal access to light.

First we compute the light environment of the tree by integrating solar data across a specified growing season as described by \cite{westling2018light}.
The integrated light environment is then raytraced through the voxelised tree point cloud and the amount of energy absorbed by each voxel is recorded.
The volume of the tree is calculated by summing the convex hulls of connected components at different height levels, using a method presented and validated by \cite{westling2020replacing}. The total light absorbed by the tree is computed by summing the light absorbed by each voxel.
Given the tree is represented by $N_v$ voxels, each with a quantity of light absorbed $L_i$, we then calculate, for each voxel, the percentage of full light absorbed by that voxel:

\begin{equation}
    p_i = \frac{L_i}{\max\limits_{j=1:N_v}L_j}
\end{equation}

\begin{figure}[ht]
    \centering
    \includegraphics[width=\columnwidth]{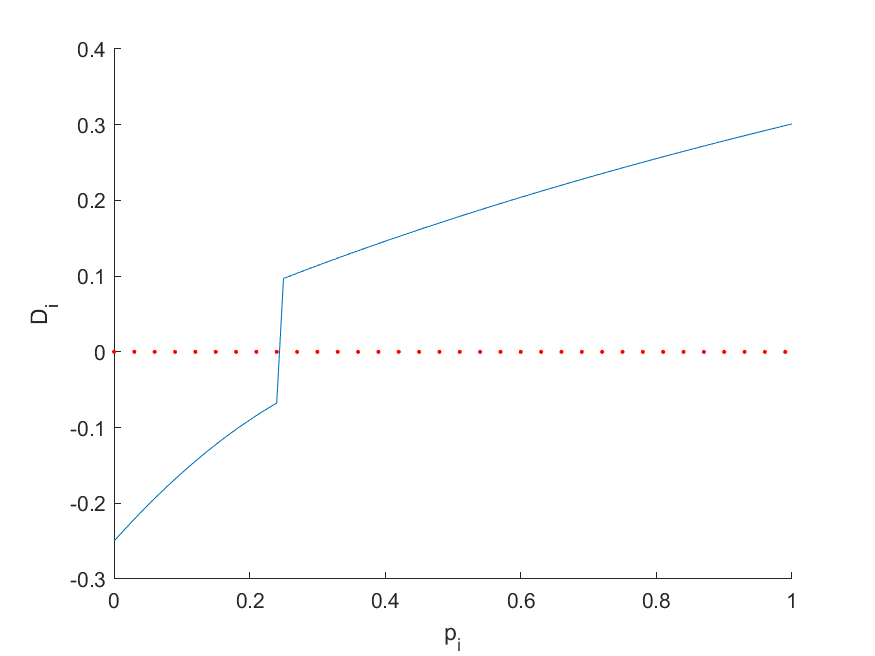}
    \caption{Response function for mapping P value to D score.  This function is designed to penalise values under 0.25 and avoid overly rewarding voxels above 0.25.  The red line denotes y=0}
    \label{fig:d-score-graph}
\end{figure}

We then calculate the "distribution score" D of the tree in order to penalise voxels under 25\%, following the target distribution presented in Figure~\ref{fig:d-score-graph}:

\begin{equation}
    D = \frac{1}{N_v} \sum\limits_{i=1:N_v}{\begin{cases}
            -(0.5 - p_i)^2, & \text{if } x\leq 0.25 \\
            log(p_i + 1),   & \text{otherwise}
        \end{cases}}
\end{equation}

Since our scoring function is intended for comparative rather than predictive purposes, we developed it to be a relative score and not an absolute one.
With that assumption, we normalise the volume and total light absorbed using the maximum value of each within the data set in question.
This prevents us from claiming if a tree is "good" or not in isolation, but allows us to judge if one tree is better than another, or whether a tree improves following a change.

\begin{equation}
    \tilde{V} = \frac{V_i}{\sum\limits_{i=1:N_{trees}}V_i} ;
    \tilde{L} = \frac{L_i}{\sum\limits_{i=1:N_{trees}}L_i},
\end{equation}
where $V_i$ is the volume of tree $i$ and $L_i$ is the sum of light absorbed by each voxel in tree $i$.

Finally, we calculate the final score $S$ for a given tree as the linear combination of the three components, to allow tuning of the importance of each.

\begin{equation}
    S = \alpha D + \beta \tilde{V} + \gamma \tilde{L}
    \label{eq:score}
\end{equation}

We calculated the values of $\alpha$, $\beta$ and $\gamma$ experimentally, using the reported yields of trees for which we have point cloud scans.
Fruit tree yield is obviously a complex value, dependent on a wide variety of factors.  However, the results presented by \cite{Orn2016} and the insights of \cite{stassen1999results} suggest it should correlate with an effective method for scoring the light distribution of trees.
For avocado trees, we had yield data for 18 datum trees in one year.
We also had yield data reporting from a mango orchard intensification trial on the Walkamin Research Station, operated by the Queensland Government Department of Agriculture and Fisheries (DAF), comprised of 270 datum trees grown using different training methods and densities, leading to a variety of tree shapes.
There were 15 independent combinations of independent variables and 6 replications of each.  When evaluating coefficients, we took the average yield and fruit weight for the 3 trees in each of these 90 experiments to suppress outliers.

Though our yield data represented different fruit trees which are different in size, shape and age, we found a set of coefficients with a similar performance for both, shown in Table~\ref{tab:coefficients}.

\begin{table}[ht]
    \centering
    \begin{tabular}{ll}
        Coefficient & Value \\
        $\alpha$    & 1.6   \\
        $\beta$     & 0.8   \\
        $\gamma$    & 0.3
    \end{tabular}
    \caption{Coefficients experimentally chosen for equation\ref{eq:score}}
\end{table}
\label{tab:coefficients}

\subsection{Pruning effect simulator}
\label{sec:method-pruning}
\begin{figure}[ht]
    \centering
    \includegraphics[width=\columnwidth]{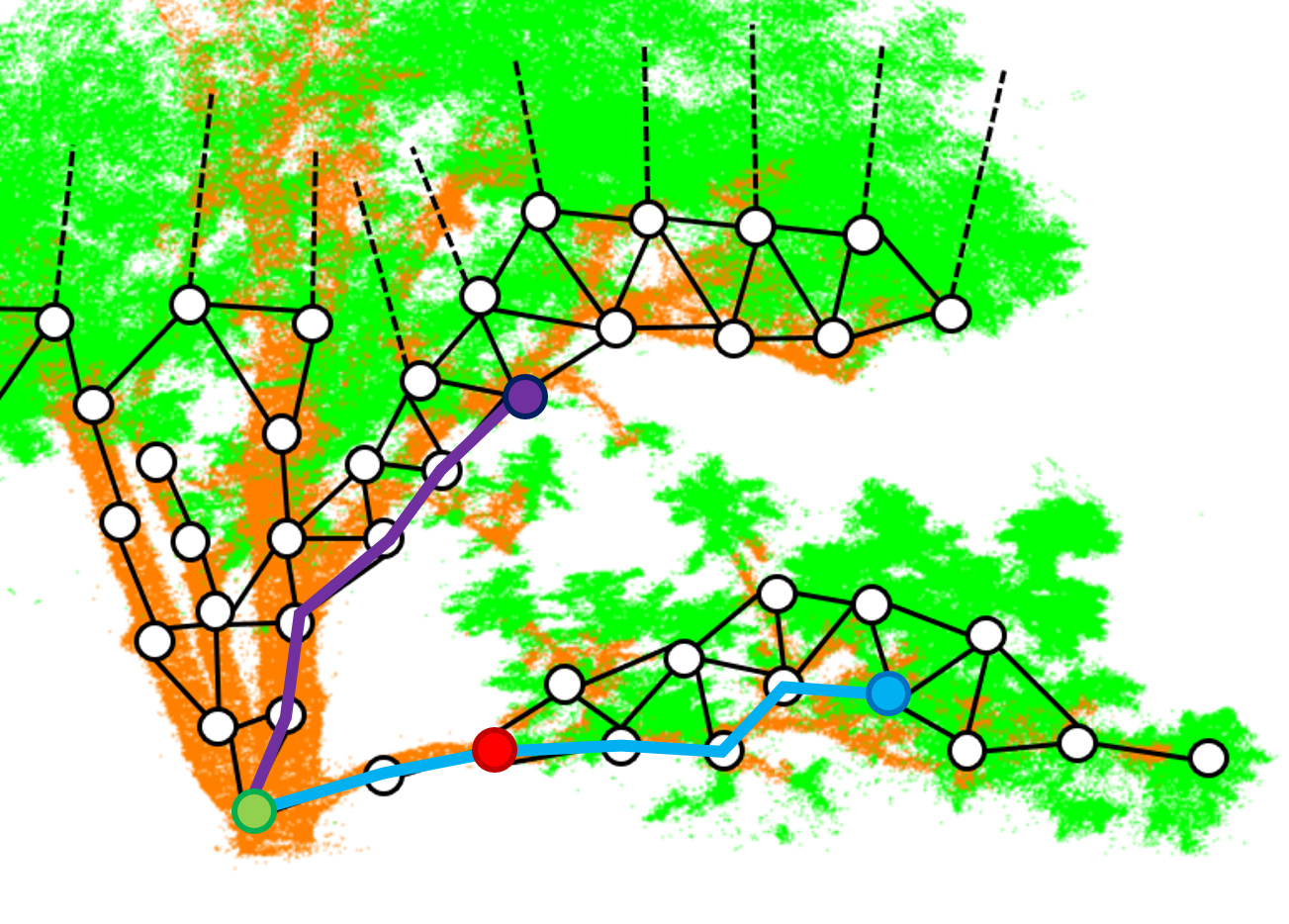}
    \caption{
        Illustration of the pruning simulation method.  An approximation of a generated graph is overlaid on the point cloud, with the node marked in green as the trunk point and the node marked in red as one of the cut points.  In this example, the purple node will be kept as it has a clear path to the trunk, while the blue node will be pruned since its path to the trunk passes through the cut point.
    }
    \label{fig:graph-method}
\end{figure}

In order to suggest cuts which improve the score of a tree, we must simulate the effect of a particular prune on the structure of the tree.
We did this using a method presented by \cite{westling2020graph} and illustrated in Figure~\ref{fig:graph-method}.
Each point cloud was voxelised to normalise matter density and speed up the processing, then a graph was created taking the mean location of the points in each voxel as the nodes, and creating edges between each node and its neighbours within a given radius.
Given a specified cut point, we marked the node at that point and all nodes within a certain radius as part of the cut.
The shortest path from each node to the trunk was then traced using A*, and each node whose path includes the cut nodes is taken as pruned, with this classification propagated to all points within that voxel.

Our real data for pruning was noisy, with several operations taking place at once and no clear definition as to where cuts were taking place.
Instead, we tested our pruning simulator by using the "SimTreeLS" LiDAR simulator presented by \cite{westling2020simtreels} to generate point clouds for which we can perfectly define removal of matter.
Three unique tree models were generated to roughly match the structural characteristics of avocado trees, and then 4 cut points were specified manually for each tree, chosen to remove one major limb with each cut.
The matter was removed from the tree in mesh form, where each leaf and limb was visible and no noise was present, allowing the matter removal ground truth to be generated without human error.
Different tree candidates were then generated, whereby each tree was generated with between 1 and 4 cuts.
Stands of 3 trees each were generated by selecting one candidate from each unique tree and placing the trees atop a ground plane in a randomised order at certain inter-tree spacings, from 3m to 8m.  8 replicates were generated at each chosen value for spacing, and these 32 stands formed our ground-truth pruned dataset $G$.
For each stand in $G$, we generated an unpruned reference stand using the same tree order and spacing, but a tree model with no cuts, to form our reference dataset $R$.
We then used SimTreeLS to "scan" all the data with a virtual LiDAR, including occlusion and sensor noise, to generate the scanned sets $L_G$ and $L_R$.
For each tree in $L_R$, we computed the pruned matter using the known cut points and the method described above to generate the pruned output set $L_P$.
We then compared the pruned trees from $L_P$ to the scanned ground truth set $L_G$, and labelled all points according to whether they should have been removed or not.  True positive points were those which were removed in $L_P$ and were not present in $L_G$, and so on.
We evaluated the success of the pruned matter detection by calculating the F1 score of the points thus labelled.

\subsection{Pruning suggestions}

\begin{figure}[ht]
    \centering
    \includegraphics[width=\columnwidth]{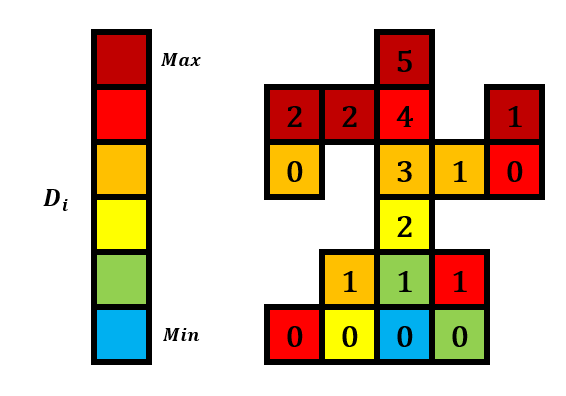}
    \caption{Illustration of process for identifying areas of the tree which should be removed to improve access to light.  Voxels are scored by counting the number of voxels below them with lower $D_i$ scores.  Viewing in colour recommended.}
    \label{fig:sum-of-count}
\end{figure}

Using our scoring function and pruning effect simulator, we developed a procedure for suggesting which limbs of the tree should be pruned to open up the canopy for more light.
We generate a number of candidate cut locations, simulate the pruning effect, and re-score the tree in each new structure.
Since thre tree score is based on raytracing, which can be time consuming, we identify candidate points by estimating parts of the tree which contribute towards the global score $S$.
To provide the necessary metadata for this, we first calculate the light distribution score $D_i$ for each voxel as described in Section~\ref{sec:method-scoring}.
As when simulating pruning effects, we then generate a graph by connecting voxels to their neighbours.
We add a "shade score" to each voxel in the graph, as illustrated in Figure~\ref{fig:sum-of-count} by counting the number of voxels directly below it with a lower $D_i$ value.
This was designed to identify voxels which were shading others, with the assumption that removing these voxels will open up the centre of the tree to more light.

\begin{figure}[ht]
    \centering
    \includegraphics[width=\columnwidth]{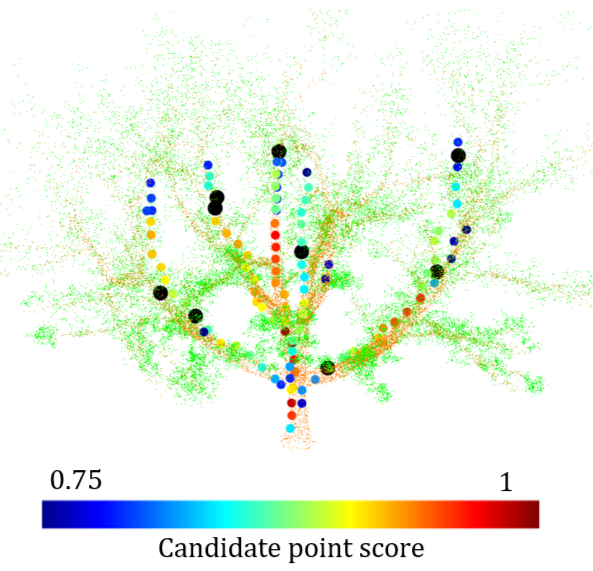}
    \caption{Illustration of pruning point suggestion.  Points coloured from blue to red (absolute values unimportant) are candidate points for pruning, namely those nodes in the graph with the highest $j_i$ score.  The black points are the final suggested cut points for this tree.}
    \label{fig:suggestions}
\end{figure}

The paths from each node to the trunk of the tree are then generated using A*.
For each node, we sum the number of times it is part of the path to a node with a high shade score.
Therefore, pruning nodes with a high sum would result in the removal of many shading nodes.
However, since this approach overwhelmingly favours nodes along the major arteries of the tree (since all paths go through them), we divide this score by the proportion of path length between the current node and the endpoint of the path and call the resultant score $j_i$.
We then select candidate points as those in the highest 5th percentile of this score.
Candidate points are then down-sampled based on distance to avoid cut points in the same location, and the $k$ highest scoring points are selected as our pruning set $N_p$.
An output of this process, as well as the scored candidate points, is shown in Figure~\ref{fig:suggestions}.

For each point in $N_p$, we simulate the effect of pruning at that point on the tree and then re-score the tree by raytracing over the new point cloud, allowing comparisons between the original score and the new scores.

\begin{figure}[h!]
    \centering
    \includegraphics[width=\columnwidth]{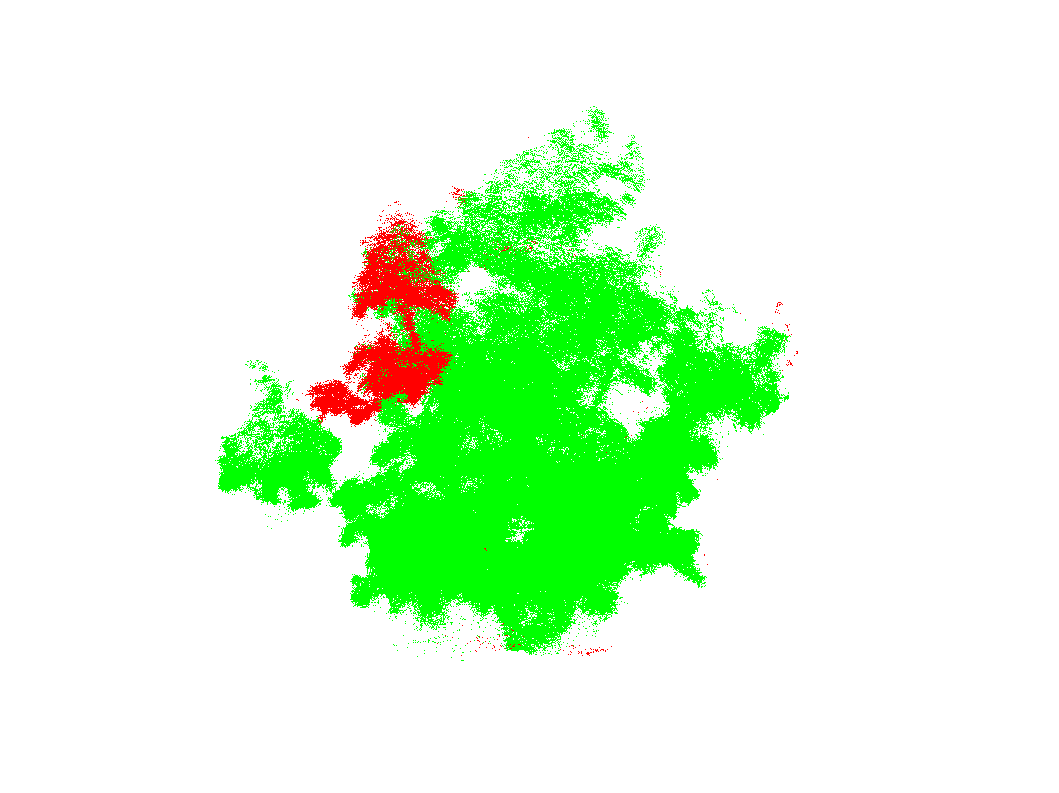}
    \caption{Visualisation of the manual pruning applied to a particular avocado tree (row-55s-tree-15e).  Points in red were removed in the selective limb removal stage of pruning.}
    \label{fig:vis-pruned-r55t15e}
\end{figure}

To validate this process, we run pruning suggestion on a virtual tree generated using SimTreeLS as well as a real tree manually pruned by commercial experts on Simpsons Farms.
The real avocado tree used was scanned following conservative application of selective limb removal, with the result showed in Figure~\ref{fig:vis-pruned-r55t15e}.

For both real and virtual data, we suggest a set $N_p$ with a small $k$ value (less than 10), and also present results when all points in $N_p$ are used rather than any individual one.

\section{Results}
\label{sec:results}

\subsection{Scoring function}
Here we present comparisons between our tree scoring function and yield, to demonstrate a favourable correlation so we can take the scoring function as a reasonable target for improving the tree shape.

As mentioned in Section~\ref{sec:method-scoring}, we scored two sets of fruit tree point clouds, one of mature avocado trees and one of young mango trees.
When evaluating the mango trees, we averaged trees within the same replicate and growing style to account for outliers, although with the avocado trees we did not have sufficient data points to do this.

\begin{figure}[ht]
    \centering
    \includegraphics[width=\columnwidth]{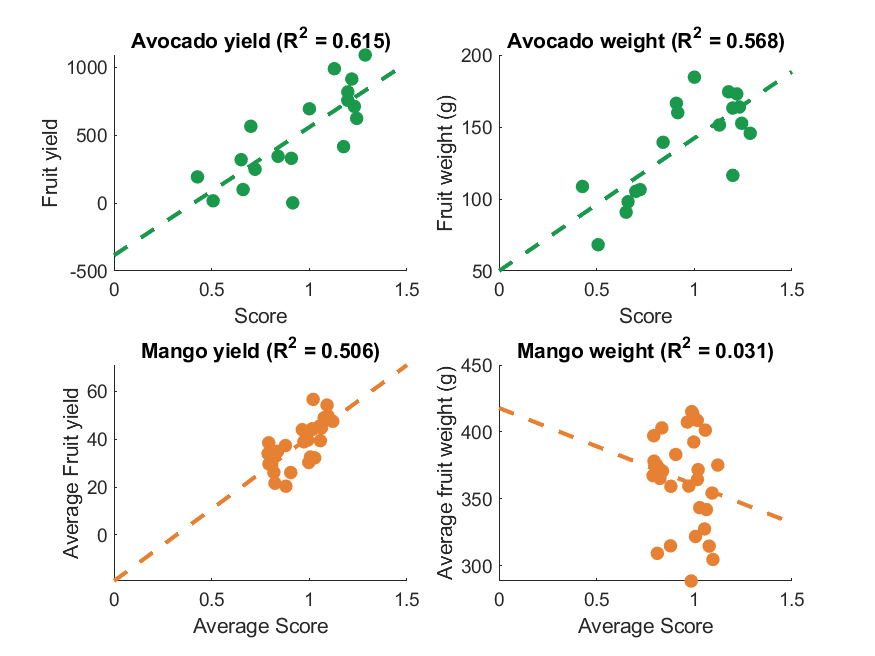}
    \caption{Comparison of scoring function as a correlator for fruit tree yield in avocado and mango trees.  Here, "yield" is the total fruit count per tree, and "weight" is the average fruit weight in grams.  In mango trees, we averaged the results for trees within the same replicate block, density and growing style to account for outliers.  In avocados, we take each of our 18 data trees individually.}
    \label{fig:score-yield}
\end{figure}

The results of both comparisons, using the coefficient values presented earlier, are shown in Figure~\ref{fig:score-yield}.


\subsection{Pruning effect simulator}

\begin{figure}[ht]
    \centering
    \includegraphics[width=\columnwidth]{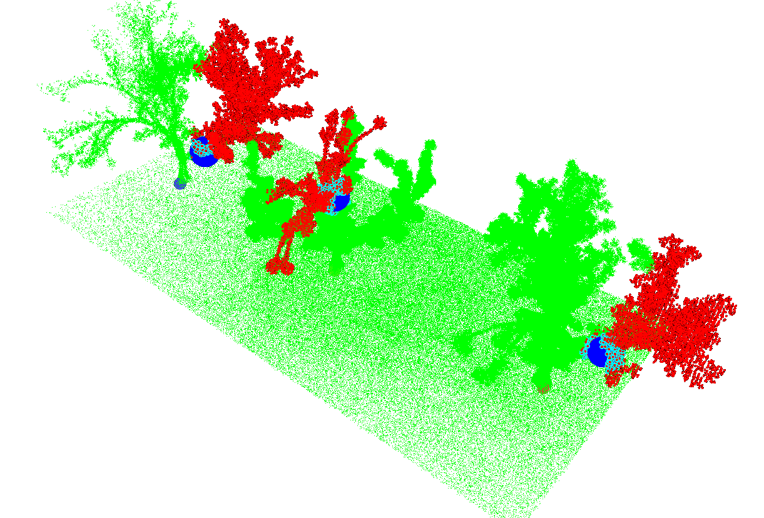}
    \caption{Example output of successful pruning matter simulation.  Large blue points are the original cut location, while cyan points are parts of the point cloud identified as part of the cut branch.  Red points are those estimated to be removed by the pruning operation.}
    \label{fig:results-qual-cuts}
\end{figure}

As described in Section~\ref{sec:method-pruning}, we calculated the F1 score of simulated matter removal for 137 trees, each with a randomly chosen number of cut points and using varying inter-tree spacing.
Figure~\ref{fig:results-qual-cuts} shows a qualitative example of the output from the pruning operation.

\begin{figure}
    \centering
    \includegraphics[width=\columnwidth]{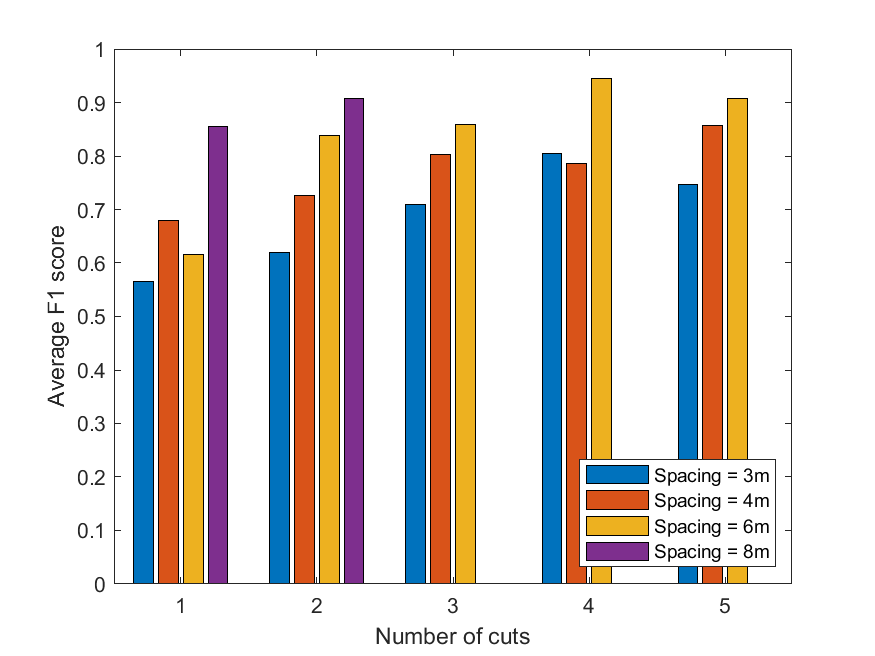}
    \caption{Results of pruning simulated trees at various tree spacings, with a randomly selected number of cut points per instance.  Scores presented are the average F1 score for all instances per spacing and number of cuts.}
    \label{fig:results-pruning-cuts}
\end{figure}

The aggregated quantitative results by number of cuts are presented in Figure~\ref{fig:results-pruning-cuts}.

\subsection{Pruning suggestions}

Tables \ref{tab:suggest-virtual} and \ref{tab:suggest-real} present the results for suggestions on a virtual and real tree respectively.
For each tree, seven points are suggested, and the score after pruning is reported.
We present the change in the overall score as well as the change in the light distribution $D$ score.
For the real tree we also present the result following manual commercial limb removal.

\begin{table}[ht]
    \begin{tabularx}{\columnwidth}{X|rrrrrr}
        Cut  & \multicolumn{1}{l}{$D$} & \multicolumn{1}{l}{$\tilde{V}$} & \multicolumn{1}{l}{$\tilde{L}$} & \multicolumn{1}{l}{S} & \multicolumn{1}{l}{$D$ change} & \multicolumn{1}{l}{S change} \\ \hline
        None & 0.269                   & 1.000                           & 1.000                           & 1.530                 & 0.00\%                         & 0.00\%                       \\
        A    & 0.277                   & 0.978                           & 0.969                           & 1.516                 & 3.12\%                         & -0.91\%                      \\
        B    & 0.282                   & 0.958                           & 0.947                           & 1.502                 & 4.95\%                         & -1.84\%                      \\
        C    & 0.297                   & 0.568                           & 0.595                           & 1.108                 & 10.36\%                        & -27.62\%                     \\
        D    & 0.301                   & 0.510                           & 0.537                           & 1.052                 & 12.11\%                        & -31.27\%                     \\
        E    & 0.282                   & 0.853                           & 0.931                           & 1.414                 & 4.99\%                         & -7.62\%                      \\
        F    & 0.280                   & 0.929                           & 0.905                           & 1.462                 & 4.04\%                         & -4.44\%                      \\
        G    & 0.277                   & 0.963                           & 0.940                           & 1.495                 & 2.90\%                         & -2.29\%                      \\
        H    & 0.278                   & 0.955                           & 0.942                           & 1.492                 & 3.47\%                         & -2.52\%                      \\
        I    & 0.279                   & 0.913                           & 0.904                           & 1.447                 & 3.65\%                         & -5.42\%                      \\
        All  & 0.315                   & 0.418                           & 0.420                           & 0.963                 & 16.96\%                        & -37.05\%                     \\
    \end{tabularx}
    \caption{Suggestions for virtual tree.  The $D$ score for the entire tree, as well as the normalised values for volume and total light absorbed are presented.  The final score $S$ and measured change in $D$ and $S$ are also presented.}
    \label{tab:suggest-virtual}
\end{table}

\begin{table}[]
    \begin{tabularx}{\columnwidth}{X|rrrrrr}
        Cuts   & \multicolumn{1}{l}{$D$} & \multicolumn{1}{l}{$\tilde{V}$} & \multicolumn{1}{l}{$\tilde{L}$} & \multicolumn{1}{l}{S} & \multicolumn{1}{l}{$D$ change} & \multicolumn{1}{l}{S change} \\ \hline
        None   & 0.182                   & 1.000                           & 1.000                           & 1.391                 & 0.00\%                         & 0.00\%                       \\
        A      & 0.193                   & 0.955                           & 0.967                           & 1.362                 & 6.02\%                         & -2.03\%                      \\
        B      & 0.198                   & 0.916                           & 0.898                           & 1.318                 & 8.84\%                         & -5.21\%                      \\
        C      & 0.195                   & 0.948                           & 0.975                           & 1.363                 & 7.33\%                         & -1.98\%                      \\
        D      & 0.204                   & 0.889                           & 0.896                           & 1.307                 & 12.50\%                        & -6.04\%                      \\
        E      & 0.185                   & 0.987                           & 0.963                           & 1.374                 & 1.66\%                         & -1.21\%                      \\
        F      & 0.188                   & 0.956                           & 0.931                           & 1.345                 & 3.39\%                         & -3.30\%                      \\
        G      & 0.189                   & 0.949                           & 0.996                           & 1.360                 & 3.80\%                         & -2.19\%                      \\
        H      & 0.185                   & 0.978                           & 0.982                           & 1.372                 & 1.59\%                         & -1.31\%                      \\
        I      & 0.200                   & 0.880                           & 0.895                           & 1.293                 & 10.16\%                        & -7.06\%                      \\
        All    & 0.227                   & 0.761                           & 0.736                           & 1.194                 & 25.15\%                        & -14.16\%                     \\
        Manual & 0.197                   & 0.808                           & 0.989                           & 1.259                 & 8.48\%                         & -9.49\%
    \end{tabularx}
    \caption{Suggestions for real tree.  The $D$ score for the entire tree, as well as the normalised values for volume and total light absorbed are presented.  The final score $S$ and measured change in $D$ and $S$ are also presented.}
    \label{tab:suggest-real}
\end{table}

\begin{figure}
    \centering
    \begin{subfigure}[t]{0.7\columnwidth}
        \centering
        \includegraphics[width=\textwidth]{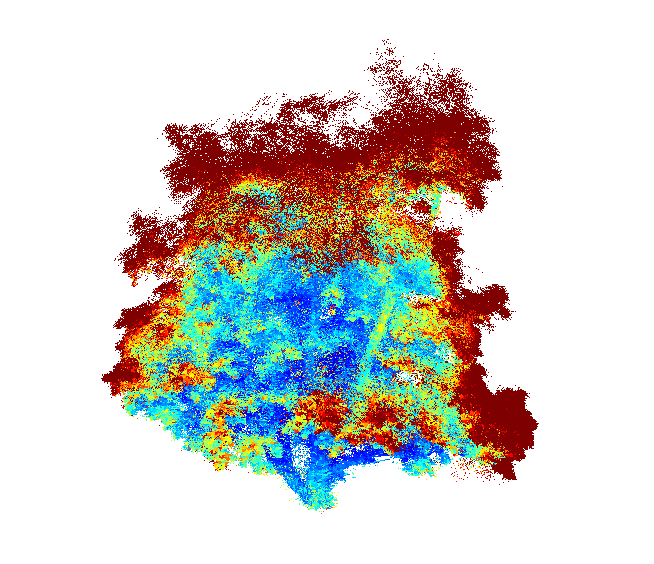}
        \caption{Unpruned}
    \end{subfigure}
    \begin{subfigure}[t]{0.7\columnwidth}
        \centering
        \includegraphics[width=\textwidth]{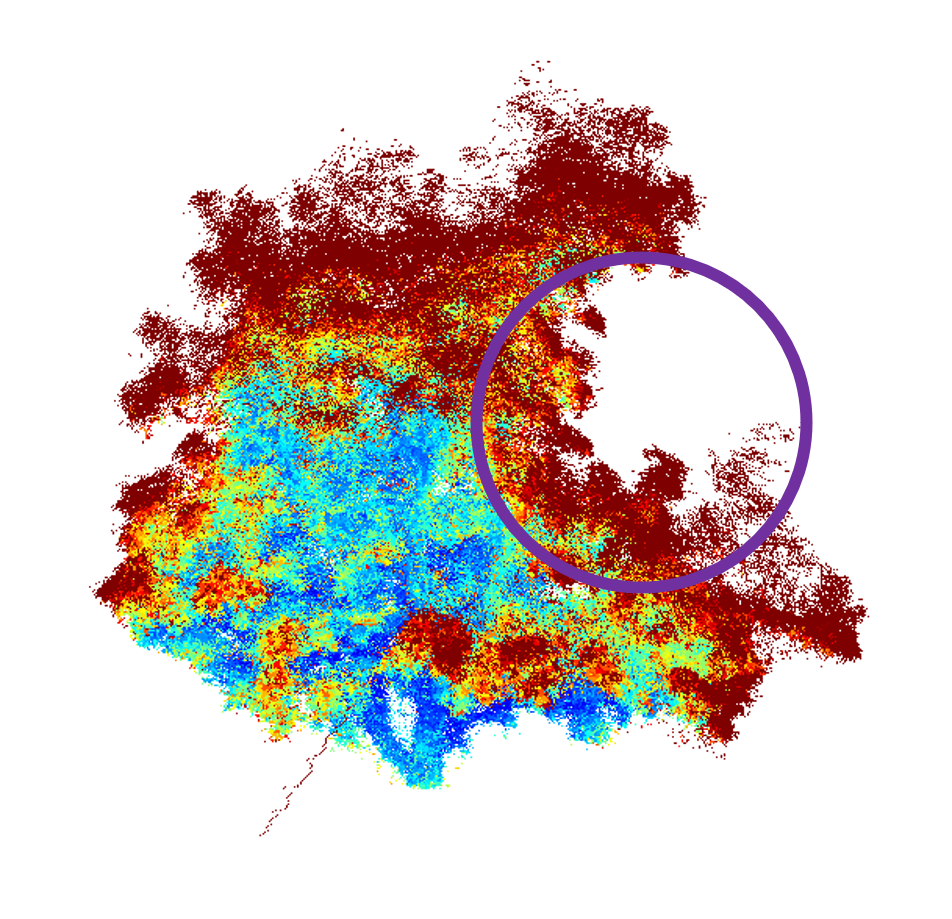}
        \caption{Pruned (manual)}
    \end{subfigure}
    \begin{subfigure}[t]{0.7\columnwidth}
        \centering
        \includegraphics[width=\textwidth]{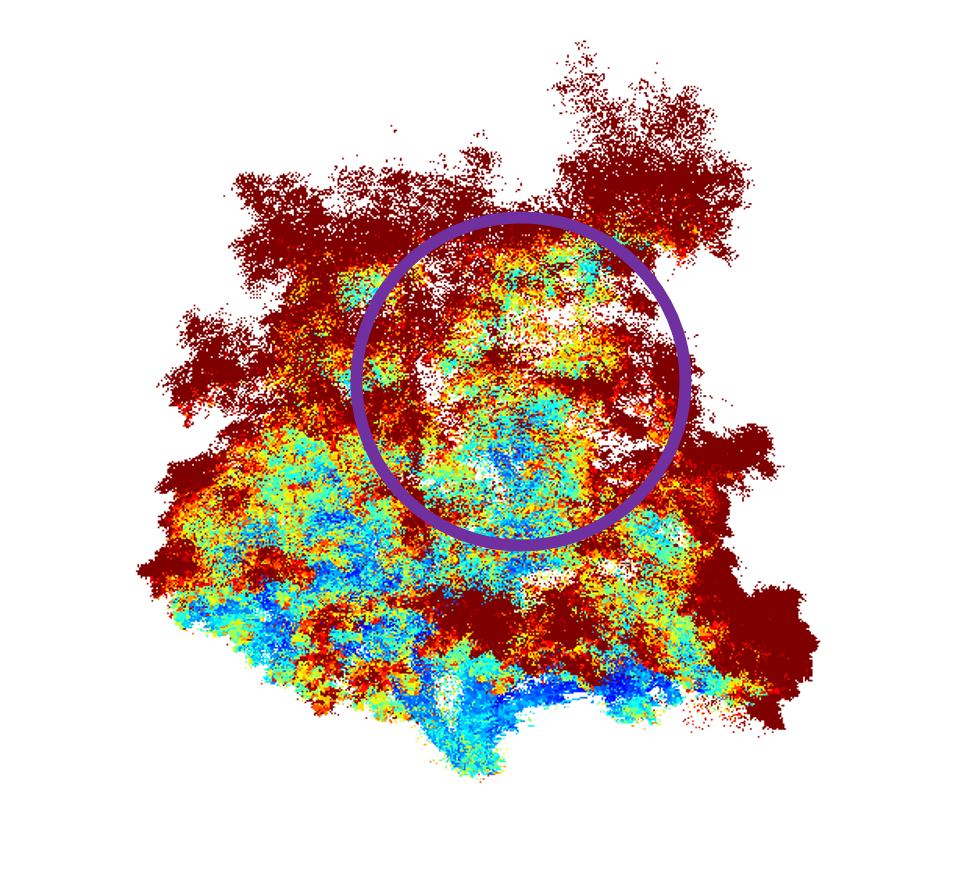}
        \caption{Pruned (ours, cut D)}
    \end{subfigure}
    \begin{subfigure}[t]{0.65\columnwidth}
        \centering
        \includegraphics[width=\textwidth]{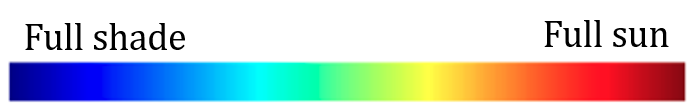}
    \end{subfigure}
    \caption{Visualisation of pruning on real tree.  Points are coloured by the amount of light absorbed during a year, and the purple circle illustrates where a limb was removed.  The "manual" point cloud is a real LiDAR capture after pruning, while our "cut D" point cloud was pruned in simulation based on an autogenerated cut point.}
    \label{fig:pruning-results-qual}
\end{figure}

Figure~\ref{fig:pruning-results-qual} shows the qualitative effect of pruning on the real avocado tree,  both with the manual commercial decision and our pruning decision, specifically cut D.

\section{Discussion}
\label{sec:discussion}

The presented framework can, given a LiDAR-scanned point cloud of a tree, suggest a number of pruning cuts for selective limb removal and then report the resulting improvement according to a defined scoring function.

Figure~\ref{fig:score-yield} presents a comparison between $S$ scores and yield characteristics.
While the correlations are far from conclusive, and fruit weight for mangoes in particular has no discernible correlation, there is evidence of a link between the score and the yield.
\cite{Orn2016} suggested that the volume of the tree is the most important property for predicting fruit count, while the total light absorbed is better for the average fruit weight.
Considering the observations of \cite{stassen1999results} and \cite{marini2020physiology}, the $D$ score is likely to be related to the sugar and oil content of the fruit produced in different parts of the canopy, with higher $D$ scores correlating with better quality fruit in more parts of the canopy.
However, that is only speculation until it can be experimentally verified.
Due to the complexity of judging tree quality and growth, our scoring function and coefficients, reflected in the yield graphs presented, are merely a starting point for the evolution of our pruning suggestion framework, to be customised depending on application.


The results presented in Figure~\ref{fig:results-pruning-cuts} shows that the pruning simulator generally works well, with an average F1 score of 0.78.
It can be observed that the results were generally better when the spacing is larger, as could be expected.  When the tree spacing is only 3m, there is significant overlap between adjacent trees, though the graph-based operation used is designed to perform reasonably well in that context.
The results also suggest the method performs slightly better with more cuts rather than fewer.  This could be caused by overlapping predictions on branches which are close together.

Results shown in Table~\ref{tab:suggest-real} demonstrate how our total suggestion procedure performs on a real avocado tree for which we also have the actual commercial pruning decision.
One thing to note is that the change in total $S$ score is always negative to varying degrees.
This is likely due to the importance of the tree volume and total light absorbed, which are always reduced when matter is removed.
The premise of pruning is that some sacrifice in volume is required to promote regrowth and improve light distribution, though this may be a deficit in our selection of coefficients.
Since we selected these by comparing to yield, the importance of light distribution (which has a smaller impact on yield than volume, as noted by \cite{Orn2016}) is reduced.
Under the assumption that improved light distribution has other benefits, we also presented the change in $D$ score as independent from the volume $V$ and total light $L$, and here we see the inverse effect - for all of our suggested cut points, as well as the manual limb removal, the D score improves after pruning.
In the case of three of our suggestions (B, D and I), the tree ends up with a higher $D$ score after pruning with a higher total score(due to less matter removed) when compared to the manually selected limb.
Cut D in particular increases $D$ by 4\% and $S$ by 3.45\% over the manual cut, and the result of this cut is shown in Figure~\ref{fig:pruning-results-qual}.
While the figure is difficult to grasp in two dimensions, our method qualitatively opens up the canopy more than the manual cut, as can be seen by the reduction in dark blue points (which tend to be shaded during the year) and the scattering of more red and orange points throughout the lower canopy.
This improved distribution comes without much physical volume being removed.

Table \ref{tab:suggest-virtual} presents similar results for our suggestion system on a virtual, computer-generated tree.
The same patterns can be seen here, with volume leading to a consistently negative change in $S$, but with the cleaner scan generated by simulation, more significant cuts were implemented with volumes up to 50\% of the tree.

These results, in particular the ones for the real tree, suggest the potential of our procedure to suggest pruning locations which can open up the canopy more while removing less matter than current commercial experts, though of course much more experimental validation would be required to confirm this.

While existing simulation-based approaches require computer-generated trees (like those generated by \cite{White2016}) or high-quality captures (like \cite{arikapudi2015orchard}), our method was developed to work with low-quality, high-noise LiDAR captures like those produced by handheld (\cite{westling2018light,bosse2012zebedee}), mobile (\cite{underwood2016mapping}), or even aerial (\cite{wu2020suitability}) equipment.

This framework overall provides a potential starting point for developing tools for commercial yield improvements, or even automated pruning.
Naturally, the parameters used can be tweaked to fit various contexts, with the basic system being appropriate for any tree structure with a desire for uniform light distribution while balancing tree volume.
Other pruning objectives can also be included without much adjustment, for instance the height-limiting approach investigated by \cite{thorp2001pruning} to move the fruiting area closer to the ground for operational reasons.
The individual parts of the framework can also be extracted or shuffled for other purposes.
One example of this could be in developing training systems like that presented by \cite{kolmanivc2017computer}, wherein a human operator suggests a cut location (taking the place of our pruning suggestion system) and is provided feedback on how well that decision would open the canopy.
Also, the pruning effect simulation could easily be adapted to perform hedging operations rather than selective limb removal, allowing the framework to perform that task as well.

\subsection{Future work}

As mentioned earlier, the scoring function we developed is fairly simple and does not take into consideration many parameters of fruit tree quality or attractiveness.
To this end, further experimentation should be undertaken to validate in particular that the $D$ score component does serve as a predictor for higher quality fruits.  This would require point cloud scanning of fruit trees, paired with a detailed survey of the yield to capture fruit quality, size and sugar/oil content in different parts of the canopy.
Other parameters can also be derived from the point cloud to add to the $S$ score.  One example could be using methods like those presented by \cite{vicari2019leaf} and \cite{westling2020graph} to segment the leaf and trunk matter, to include properties like the woody-to-total area ratio described by \cite{Ma2016}.

The suggestion method itself could also be further developed using more sophisticated gradient descent approaches, or a more expansive optimisation technique.
The approach suggested here is also simplistic in its computation of shade score, assuming that each voxel only shades those directly below it, while the truth could be more accurately modelled as a cone of influence, with parameters inferred by the integrated sky.

With further development, this framework could also include regrowth simulation to more accurately model the long-term effects of pruning, as presented by \cite{strnad2020framework}.
This would require some careful interpolation, given that methods like these typically require a higher fidelity tree model than can be captured in reasonable time using currently available LiDAR technology.

\section{Conclusion}
\label{sec:conclusion}
We have presented a framework for suggesting strategies for selective limb removal pruning on fruit trees, with a specific focus on avocados and mangoes.
Our results show correlation between the scoring function we developed and yield, and we are able to successfully simulate the removal of plant matter given specific cut points.
By dynamically suggesting cut points in order to remove tree matter which is casting shade on parts of the tree, we also presented a method for suggesting cuts which improved the light distribution through the canopy by up to 25.15\%.
The steps in this framework can be further developed to provide a pruning suggestion system with commercial applicability.

\subsubsection*{Acknowledgements}
This work is supported by the Australian Centre for Field Robotics (ACFR) at The University of Sydney.  For more information about robots and systems for agriculture at the ACFR, please visit http://sydney.edu.au/acfr/agriculture.

\nocite{Tange2011a} 

\bibliography{references}{}
\end{document}